\documentclass[letterpaper, 10 pt, conference]{ieeeconf}  

\IEEEoverridecommandlockouts                              

\usepackage{amsmath,amsfonts}
\usepackage{algorithm}
\usepackage{algpseudocode}
\usepackage{array}
\usepackage[caption=false,font=normalsize,labelfont=sf,textfont=sf]{subfig}
\usepackage{textcomp}
\usepackage{xcolor}
\usepackage{stfloats}
\usepackage{hyperref}
\usepackage{url}
\usepackage{booktabs}
\usepackage{verbatim}
\usepackage{graphicx}
\usepackage{optidef}
\usepackage[style=ieee,doi=false,isbn=false,url=false,eprint=false]{biblatex}

\newcommand{\R}[1]{\mathbf{R}^{#1}}

\title{\LARGE \bf
Differentiable Collision Detection for a Set of Convex Primitives
}

\author{Kevin Tracy$^{1}$, Taylor A. Howell$^{2}$, and Zachary Manchester$^{1}$
\thanks{$^{1}$Kevin Tracy and Zachary Manchester are with The Robotics Institute, Carnegie Mellon University, Pittsburgh, PA 15213, USA
        {\tt\small \{ktracy,zacm\}@cmu.edu}}%
\thanks{$^{2}$Taylor Howell is with the Department of Mechanical Engineering, Stanford University, Stanford, CA 94305, USA
        {\tt\small thowell@stanford.edu}}%
}

\begin{document}

\maketitle
\thispagestyle{empty}
\pagestyle{empty}




\begin{abstract}
Collision detection between objects is critical for simulation, control, and learning for robotic systems. However, existing collision detection routines are inherently non-differentiable, limiting their applications in gradient-based optimization tools. In this work, we propose DCOL: a fast and fully differentiable collision-detection framework that reasons about collisions between a set of composable and highly expressive convex primitive shapes. This is achieved by formulating the collision detection problem as a convex optimization problem that solves for the minimum uniform scaling applied to each primitive before they intersect.  The optimization problem is fully differentiable with respect to the configurations of each primitive and is able to return a collision detection metric and contact points on each object, agnostic of interpenetration. We demonstrate the capabilities of DCOL on a range of robotics problems from trajectory optimization and contact physics, and have made an open-source implementation available.
\end{abstract}
\section{Introduction}
Computing collisions is of great interest to the computer graphics, video game, and robotics communities. Popular algorithms for collision detection include the Gilbert, Johnson, and Keerthi (GJK) algorithm \cite{gilbert1988}, its updated variant enhanced-GJK \cite{cameron1997}, and Minkowski Portal Refinement (MPR) \cite{snethen2008,newth2013}.  For objects that have interpenetration, the Expanding Polytope Algorithm (EPA) \cite{vandenbergen2001} is used to return a metric that describes the depth of penetration between two objects. These algorithms are implemented in the widely used Flexible Collision Library (FCL) \cite{pan2012}, and are employed in most physics engines including Bullet \cite{coumans2015}, Drake \cite{tedrake2019a}, Dart \cite{lee2018}, and MuJoCo \cite{todorov2012a}. While efficient and robust, all of these algorithms are inherently non-differentiable due to their logical control flow and pivoting.  

Two methods have been proposed for calculating approximate gradients of a collision metric: The first is sample-based randomized smoothing of GJK for finite-differenced gradients \cite{montaut2022a}, and the second formulates the closest distance between spheres, capsules, planes, and boxes, as a differentiable optimization problem \cite{zimmermann2022}, similar to \cite{tracy2022}.  The approximate gradients from the first method are expensive to compute and are unable to return useful information if penetration occurs, while the second method is only able to handle a limited selection of convex primitives.

This paper introduces DCOL, a differentiable collision-detection framework that computes closest points, minimum distance, and interpenetration depth between any pair of six convex primitive shapes: polytopes, capsules, cylinders, cones, ellipsoids, and padded polygons (Fig. \ref{fig_sim}). We do this by formulating a convex optimization problem that solves for the minimum uniform scaling that must be applied to the primitives for an intersection to occur, an idea first proposed for polytopes in \cite{gilbert1994}. When primitives are not in contact, the minimum scaling for an intersection is greater than one, and when there is interpenetration between objects, the minimum scaling is less than one.  The ability to return an informative collision metric in the presence of interpenetration is a key distinction between DCOL and GJK variants.  In addition to the scaling parameter detecting a collision, the contact points on each object can be calculated from this solution as well.
\begin{figure}[t!]
\centerline{\includegraphics[width = 7.5cm]{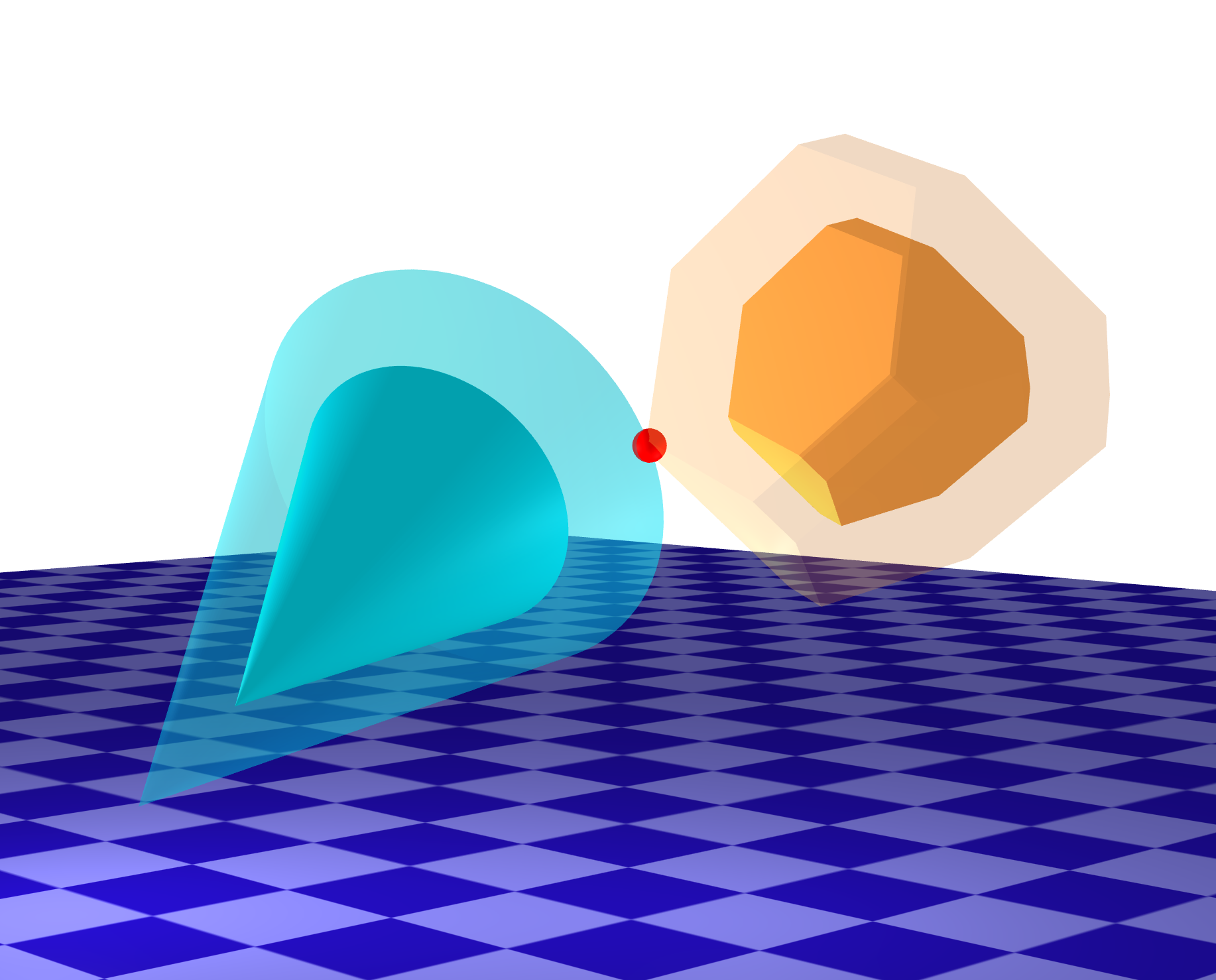}}
\caption{Collision detection between a cone and a polytope. DCOL works by solving an optimization problem for the minimum scaling of each object that produces an intersection which, in this example, is greater than one, meaning there is no collision. The scaled objects are translucent and the intersection point between these scaled objects is shown in red.}
\label{fig:conepoly}
\vspace{-10pt}
\end{figure}

The optimization problems produced by our formulation are bounded, feasible, and well-defined for all configurations of the primitives. Differentiable convex optimization allows the sensitivities of the solution to be calculated with respect to problem parameters with minimal added computation. This allows informative and smooth derivatives of the minimum scaling as well as the contact points with respect to the configurations of the primitives to be computed efficiently.

The ability to differentiate through our collision detection algorithm enables the inclusion of accurate collision information into gradient-based robotic simulation, control, and learning frameworks. We demonstrate this on several relevant robotics problems from trajectory optimization and contact physics. 

Our specific contributions in this paper are the following:
\begin{itemize}
    \item An optimization-based collision detection formulation between convex primitives that returns an informative collision metric even in the case of interpenetration
    \item Efficient differentiation of this optimization problem with respect to the configurations of each primitive 
    \item A fast and efficient open-source implementation of these algorithms built on a custom primal-dual interior-point solver
\end{itemize}

The paper proceeds by providing background on standard convex conic optimization and differentiation of conic optimization problems in Section \ref{sect:conic_opt}, a derivation of DCOL in Section \ref{sect:framework} with the corresponding constraints for each of the six convex primitives shown in Fig. \ref{fig_sim}, example use cases in trajectory optimization and contact physics in Section \ref{sect:examples}, and our conclusions in Section \ref{sect:conclusion}.

 \section{Background} \label{sect:conic_opt}
 The differentiable collision detection algorithm, DCOL, proposed in this paper is built on differentiable convex optimization. In this section, convex optimization with the relevant conic constraints is detailed, as well as a method for efficiently computing derivatives of these optimization problems with respect to problem parameters. 
 \subsection{Conic Optimization}
 DCOL formulates collision detection problems as a convex optimization problem with conic constraints \cite{boyd2004}. In standard form, these optimization problems have linear objectives and constraints of the following form:
 \begin{mini}
{x}{ c^Tx }{\label{conic_form}}{}
\addConstraint{h-Gx}{\in \mathcal{K},}
\end{mini}
where $x\in\R{n}$, $c \in \R{n}$, $G \in \R{m \times n}$, $h \in \R{m}$, and $\mathcal{K} = \mathcal{K}_1 \times \dotsm \times \mathcal{K}_N$ is a Cartesian product of $N$ proper convex cones. The optimality conditions for problem \ref{conic_form} are as follows:
\begin{align}
    c + G^Tz &= 0, \label{conic_kkt_1}\\
    h - Gx &\in \mathcal{K}, \label{conic_kkt_2}\\ 
    z &\in \mathcal{K}^*, \label{conic_kkt_3}\\ 
    (h-Gx) \circ z &= 0, \label{conic_kkt_4}
\end{align}
where a dual variable $z \in \R{m}$ is introduced, $\mathcal{K}^*$ is the dual cone, and $\circ$ is a cone product specific to each cone  \cite{vandenberghe}. 

The cones required for DCOL include the nonnegative orthant, denoted as $\mathbf{R}_+^m$, and the second-order cone, denoted as $\mathcal{Q}_m$. The nonnegative orthant contains any vector $s \in \mathbf{R}_+^m$ where $s \geq 0$, and the a second-order cone contains any vector $s \in \mathcal{Q}_m$ such that $\|s_{2:m}\|_2 \leq s_1$.

We develop a custom primal-dual interior-point solver for DCOL, with support for both of the relevant cones, based on the \textit{conelp} solver from \cite{vandenberghe}, with features taken from  \cite{domahidi2013a,nesterov1997,andersen2003,nesterov1998}.  The memory for this custom solver is entirely stack-allocated and is optimized for the small problems that DCOL creates, dramatically outperforming off-the-shelf primal-dual interior-point conic solvers like ECOS \cite{domahidi2013a} and Mosek \cite{mosekaps2014}.

\subsection{Differentiating Through a Cone Program}
Recent advances in differentiable convex optimization have enabled efficient differentiation through problems of the form \eqref{conic_form} \cite{agrawal2019,agrawal2019a,amos2019}.  Solutions to \eqref{conic_form} can be differentiated with respect to any parameters used in $c$, $G$, and $h$. 

At the core of differentiable convex optimization is the implicit function theorem. An implicit function $g:\R{a} \times \R{b} \rightarrow \R{a} $ is defined as:
\begin{align}
    g(y^*,\theta) &= 0 ,\label{ift:res}
\end{align}
for an equilibrium point $y^* \in \R{a}$, and problem parameters $\theta \in \R{b}$. Approximating \eqref{ift:res} with a first-order Taylor series results in: 
\begin{align}
    \frac{\partial g}{\partial y} \delta y + \frac{\partial g}{\partial \theta} \delta \theta &= 0 ,
\end{align}
which can be re-arranged to solve for the sensitivities of the solution with respect to the problem parameters:
\begin{align}
    \frac{\partial y}{\partial \theta} &= - \bigg( \frac{\partial g}{\partial y} \bigg)^{-1} \frac{\partial g}{\partial \theta}. \label{eq:ift}
\end{align}
By treating the optimality conditions in equations \eqref{conic_kkt_1} and \eqref{conic_kkt_4} as an implicit function at a primal-dual solution, the sensitivities of the solution with respect to the problem data can be computed. When the original optimization problem is solved using a primal-dual interior-point method as described in \cite{vandenberghe}, these derivatives can be computed after the solve without any additional matrix factorizations \cite{amos2019}.  This enables fast differentiation of conic programs that are fit for use in our differentiable collision detection algorithm.

In the case where only the gradient of the objective value $J$ with respect to the problem parameters $\theta$ is needed, the implicit function theorem is unnecessary. Instead, we need only the gradient of the Lagrangian for \eqref{conic_form}: 
\begin{align}
    \mathcal{L}(x,z,\theta) &= c(\theta)^Tx + z^T(G(\theta) x - h(\theta)),
\end{align}
where the problem matrices $c$, $h$, and $G$, are functions of the problem parameters $\theta$. Given a primal-dual solution $(x^*,z^*)$, the gradient of the objective value with respect to the problem parameters is simply the gradient of the Lagrangian with respect to these problem parameters, $ \nabla_\theta J = {\nabla_\theta \mathcal{L}(x^*,z^*,\theta)}$.
This allows for a faster computation of this specific gradient without using the implicit function theorem.
\section{The DCOL Algorithm}\label{sect:framework}
\begin{figure*}[!t]
\centering
\subfloat[]{\includegraphics[width=1.8in]{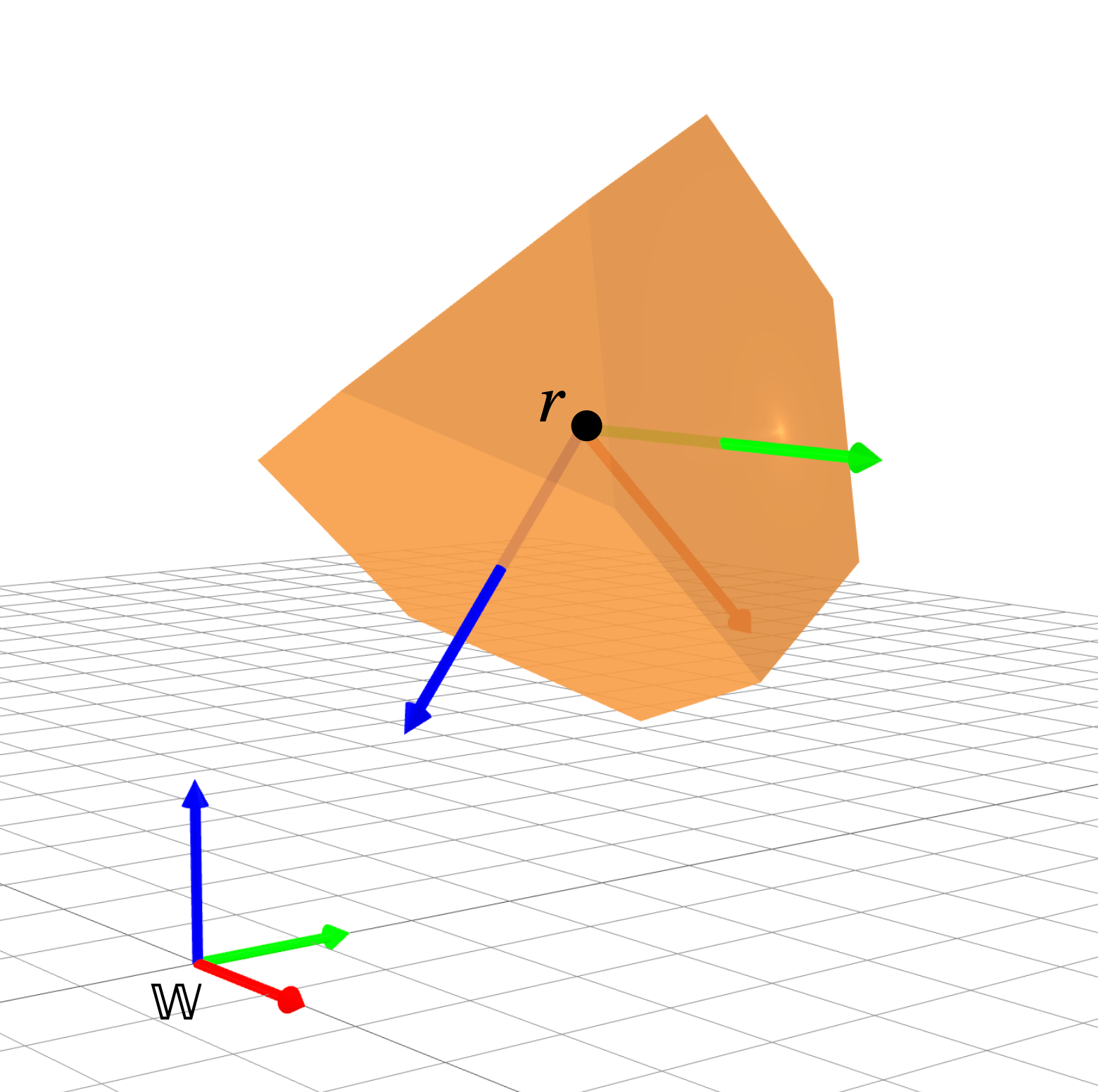}%
\label{fig_polytope}}
\hfil
\subfloat[]{\includegraphics[width=1.8in]{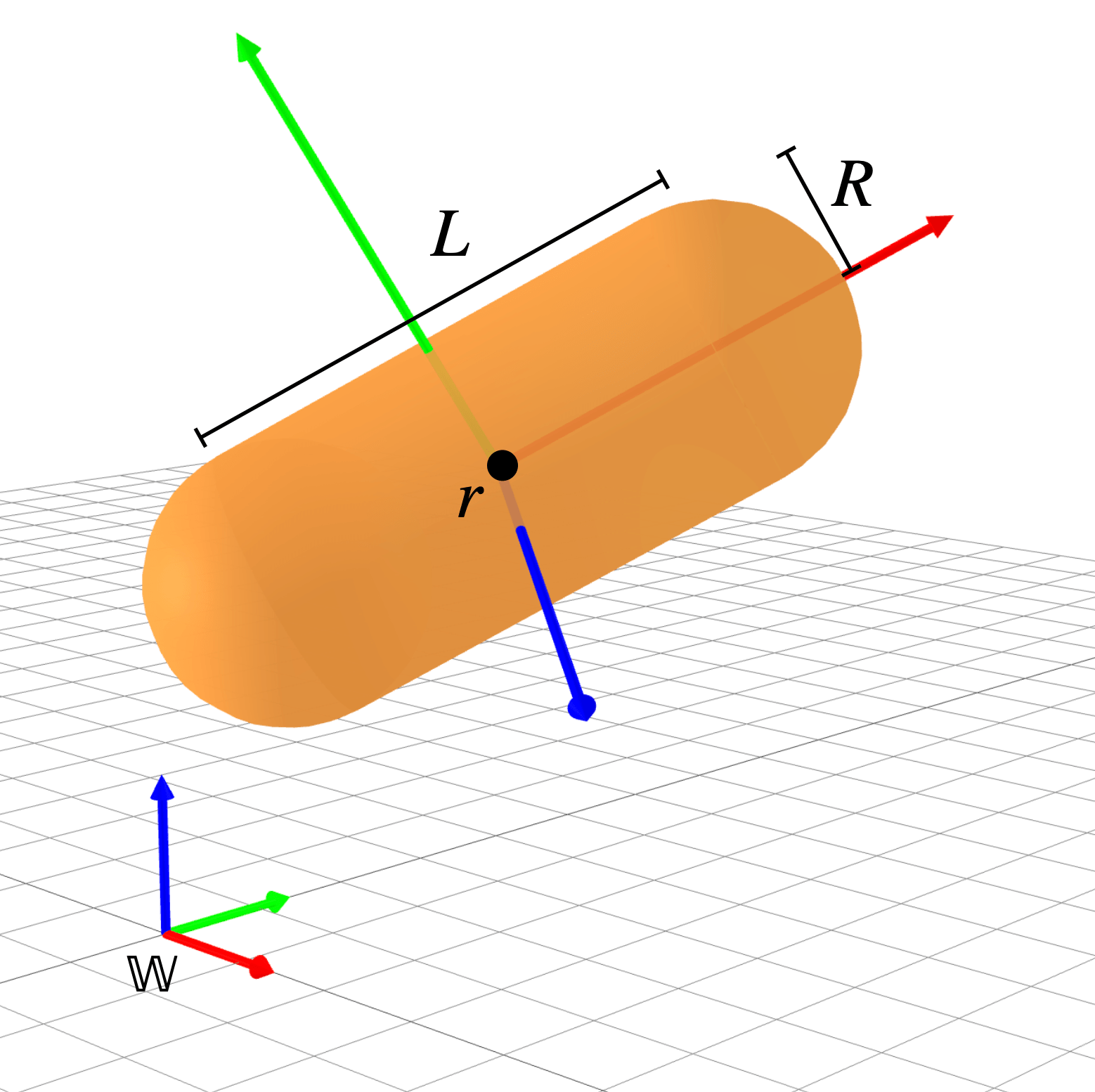}%
\label{fig_capsule}}
\hfil
\subfloat[]{\includegraphics[width=1.8in]{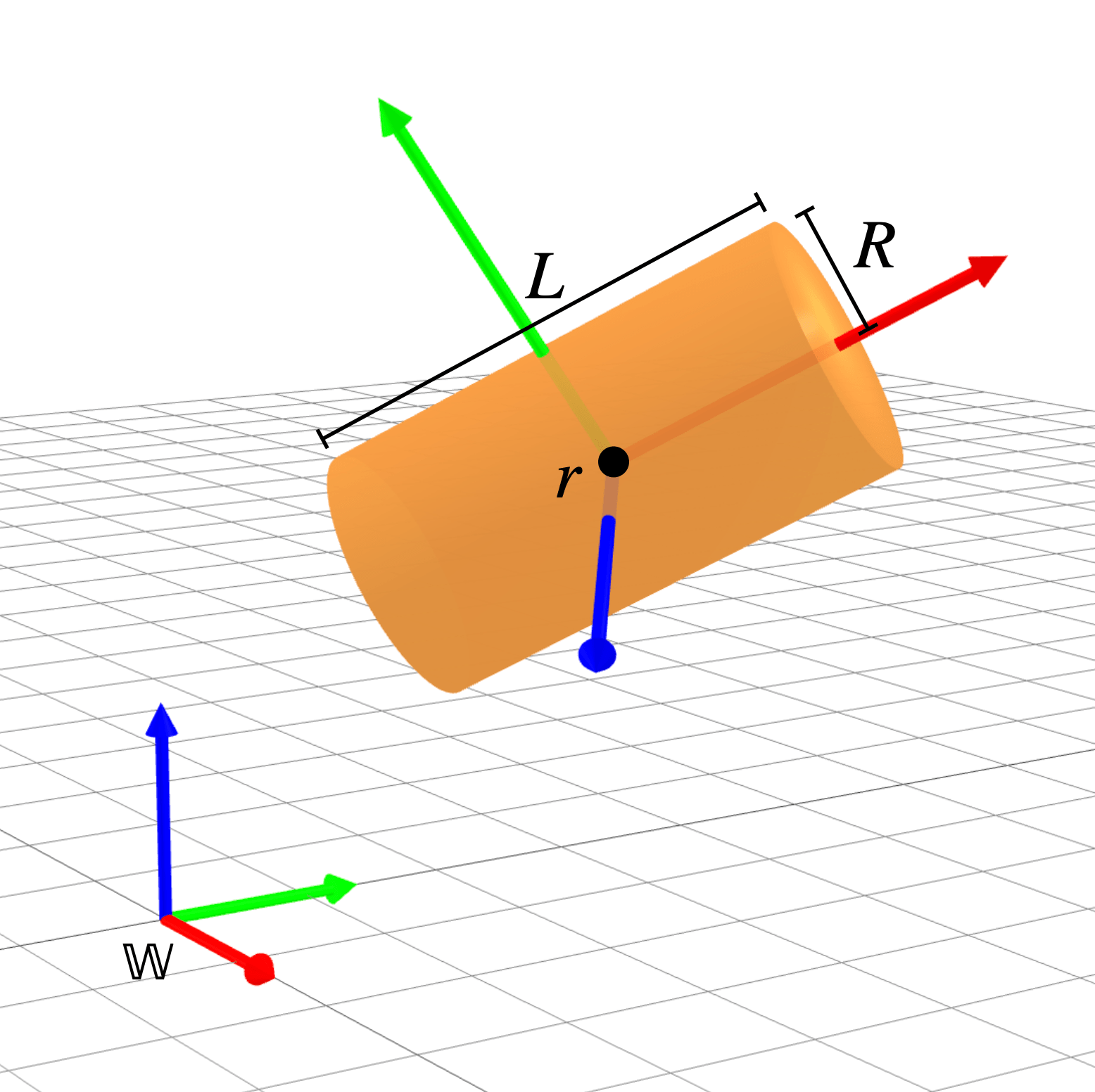}%
\label{fig_cyl}}
\hfil
\subfloat[]{\includegraphics[width=1.8in]{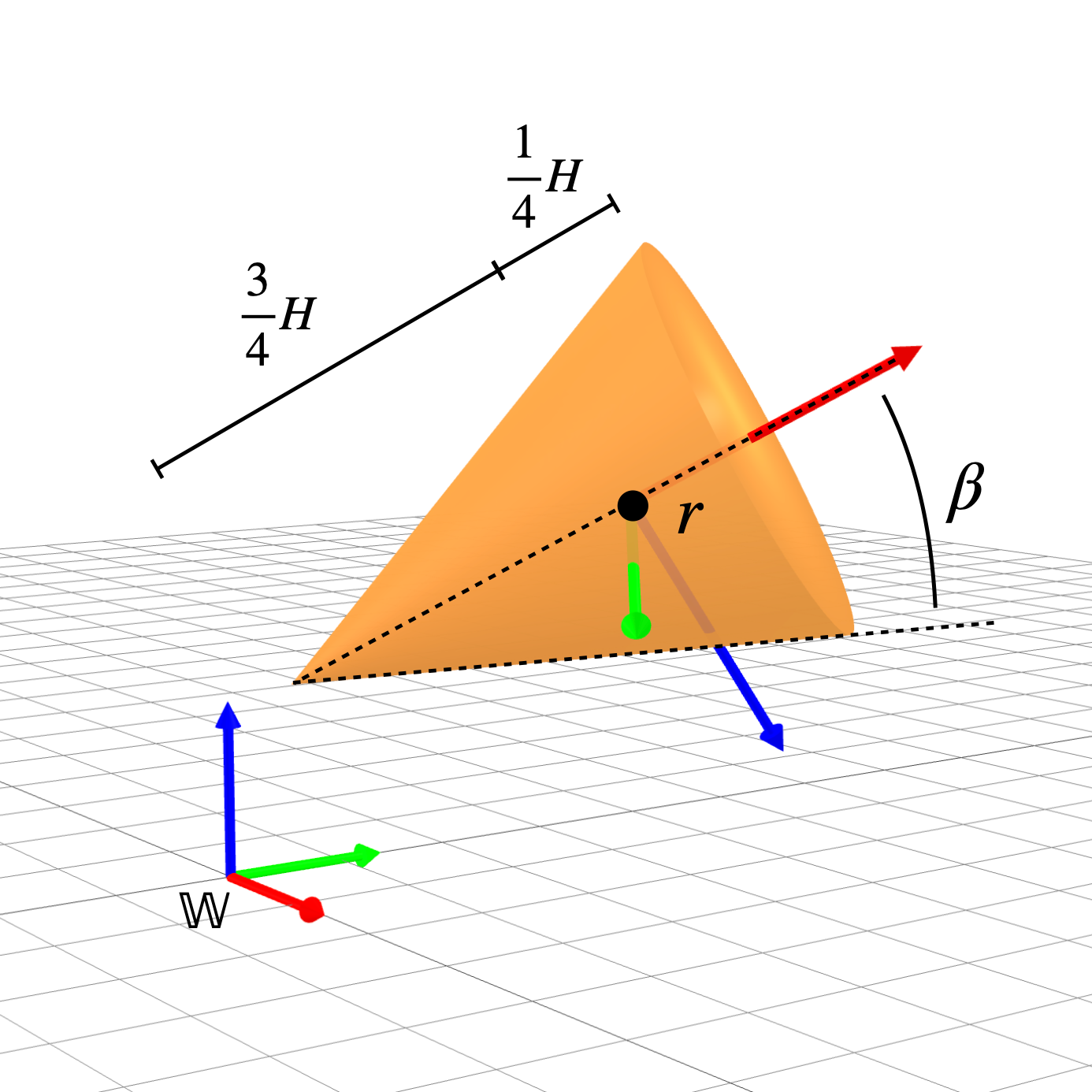}%
\label{fig_cone}}
\hfil
\subfloat[]{\includegraphics[width=1.8in]{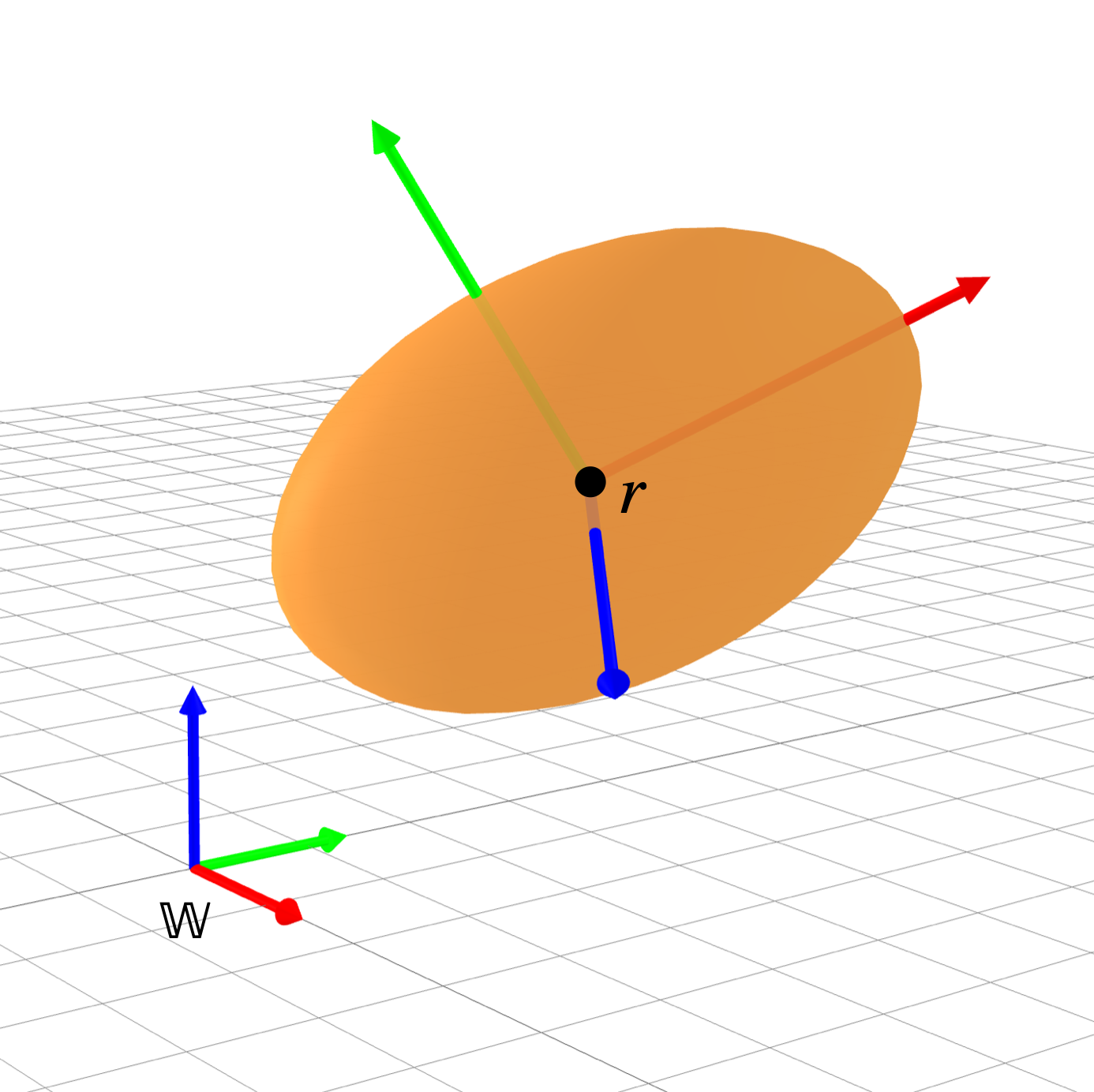}%
\label{fig_sphere}}
\hfil
\subfloat[]{\includegraphics[width=1.8in]{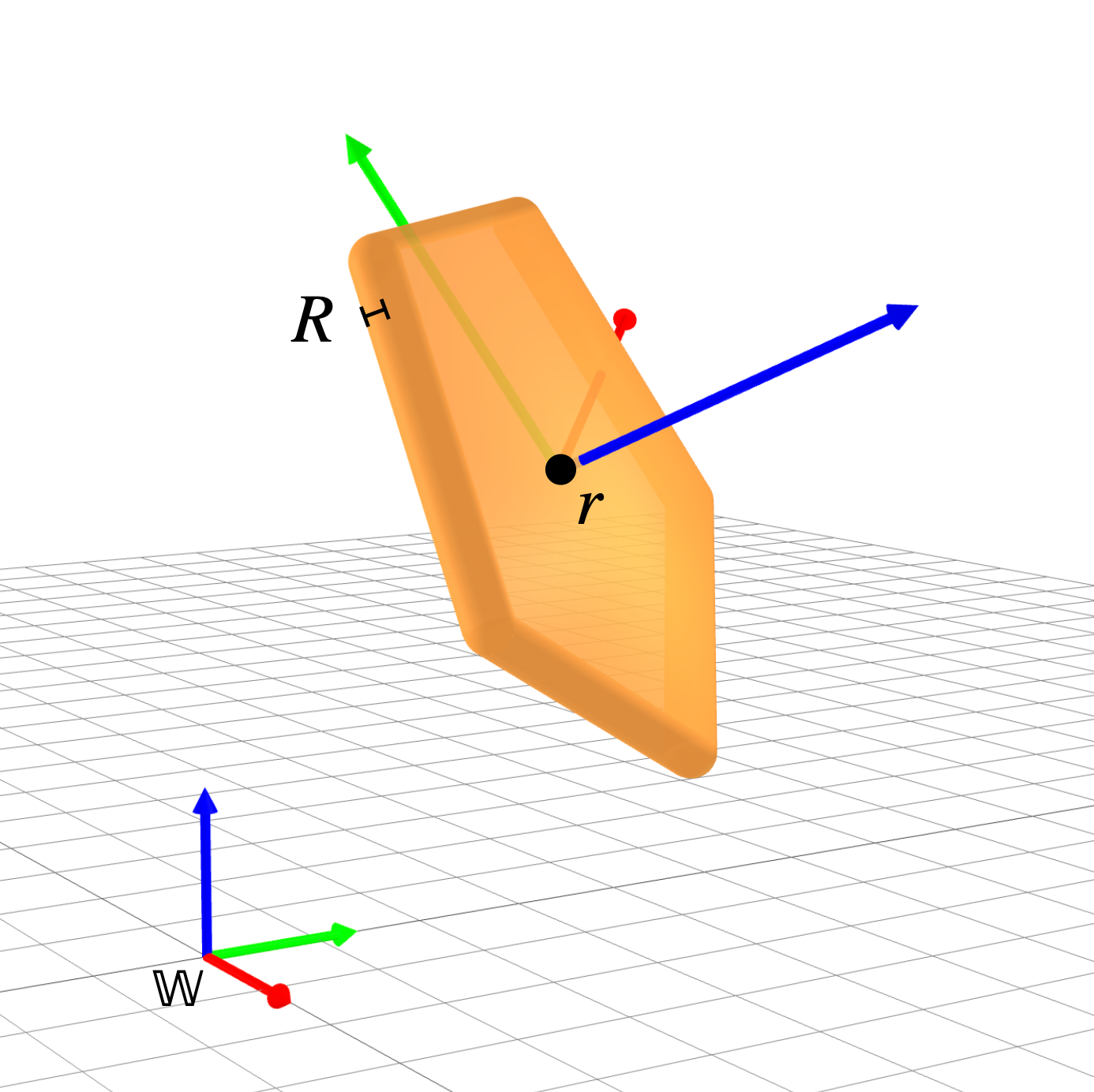}%
\label{fig_polyg}}
\caption{Geometric descriptions of the six primitive shapes that are compatible with this differentiable collision detection algorithm. These shapes include a polytope (a), capsule (b), cylinder (c), cone (d), ellipsoid (e), and padded polygon (f). Collision information including the collision status as well as the contact points can be computed between any of two of these primitives using DCOL.}
\label{fig_sim}
\end{figure*}
This section details how DCOL computes collision information between two convex primitives.  The core part of this framework is an optimization problem that solves for a minimum uniform scaling $\alpha \in \R{}$ applied to both objects that result in an intersection. In the case where there is no collision between the two objects, the minimum scaling is greater than one, and when there is interpenetration, the minimum scaling is less than one. Because of this, we find the minimum scaling $\alpha$ is a better collision metric than the closest distance between the primitives, allowing for the straightforward description of collision constraints that are agnostic of interpenetration.  All steps in the creation and solving of this optimization problem are fully differentiable, and average timing results for computing both solutions and derivatives are provided in Table \ref{speed_table} as an average over each primitive.
\begin{table}[!t]
\centering 
\caption{Average DCOL Computation Times}
\begin{tabular}{c c c c c c c}
		\toprule
	                 &
                  \textbf{polyt.} & 
                  \textbf{caps.} & 
                  \textbf{cyl.} & 
                  \textbf{cone} & 
                  \textbf{ellips.} & 
                  \textbf{polyg.} \\
		\toprule
        evaluate & 5.9 $\mu$s &  8.5 $\mu$s     & 8.4 $\mu$s     & 5.0 $\mu$s     & 6.8 $\mu$s        & 9.4 $\mu$s       \\
differentiate &  1.4 $\mu$s                        &  1.4 $\mu$s     &  1.6 $\mu$s    &  1.3 $\mu$s    &   1.3 $\mu$s      &  1.7 $\mu$s      \\
		\toprule
\end{tabular}
\label{speed_table}
\end{table}
\subsection{Optimization Problem}
Scaled convex primitives are described as a set $S(\alpha)$, which is a specific instance of the primitive scaled by some $\alpha$. A point $x\in \R{3}$ is said to be in the set $x \in S(\alpha)$ if $x$ is within the scaled primitive. This notation allows for the following formulation of the optimization problem:
 \begin{mini}
{x, \alpha}{ \alpha }{\label{dcd}}{}
\addConstraint{x}{\in \mathcal{S}_1(\alpha)}
\addConstraint{x}{\in \mathcal{S}_2(\alpha)}
\addConstraint{\alpha}{\geq 0,}
\end{mini}
where the minimum scaling $\alpha$ is computed such that $x$ is in the interior of both of the scaled primitives, making $x$ an intersection point. This optimization problem is convex, bounded, and feasible for all of the primitives described in this paper. The boundedness comes from the constraint $\alpha \geq 0$, and the guarantee of feasibility comes from the fact that each object is uniformly scaled, so that in the limit $\alpha \rightarrow \infty$ each shape will encompass the entirety of $\R{3}$, guaranteeing an intersection between objects. Another benefit to this problem formulation is that the only time the minimum scaling $\alpha = 0$ is when the origins of the two objects are coincident, in which the problem and its derivatives are still well defined.
\subsection{Primitives}
This section details the constraints that define set membership for each of the six scaled primitives.  Each object is defined with an attached body reference frame $\mathbb{B}$ with an origin $r\in\R{3}$ expressed in a world frame $\mathbb{W}$. The uniform scaling of these objects is always centered about this position $r$, and when the scaling parameter $\alpha$ is $0$, the object is simply a point centered at $r$.  The orientation of an object is defined by a rotation matrix ${}^\mathbb{W} Q {}^\mathbb{B} \in \R{3 \times 3}$ relating the world frame to the object-fixed body frame, denoted as $Q$ for shorthand.  For each primitive, the constraints are also explicitly written in standard conic form for direct inclusion in our custom conic solver in the form of \eqref{conic_form}, where $h - Gx \in \mathcal{K}$.
\subsubsection{Polytope}
A polytope is a convex shape in $\R{3}$ defined by a set of halfspace constraints, an example of which is shown in Fig. \ref{fig_polytope}. This polytope is described as the set of points $w \in \R{3}$ such that  $Aw\leq b$ for $w$ expressed in $\mathbb{B}$, where $A \in \R{m \times 3}$ and $b \in \R{m}$ represent the $m$ halfspace constraints comprising the polytope. This polytope can be scaled by $\alpha$, resulting in the following constraint for $x$ to be inside the polytope:
\begin{align}
    AQ^T (x - r) &\leq \alpha b.
\end{align}
The scaling parameter $\alpha$ scales the vector $b$, resulting in uniform scaling of all halfspace constraints and subsequent uniform scaling of the polytope. This constraint in standard form is the following: 
\begin{align}
   AQ^Tr  - \begin{bmatrix} A Q^T & -b \end{bmatrix} \begin{bmatrix} x \\ \alpha \end{bmatrix} &\in \mathbf{R}_+ .
\end{align}
\subsubsection{Capsule}
A capsule can be defined by the set of points within some radius $R$ of a line segment, as shown in Fig. \ref{fig_capsule}.  This internal line segment is along the $x$ axis of the attached reference frame $\mathbb{B}$, and the end points of this line segment are some distance $L$ apart. The scaled constraints for this primitive are that the point $x$ must be within a scaled radius of the line segment, where the distance of the endpoints of the line segment from $r$ is also scaled: 
\begin{align}
    \| x - (r + \gamma \hat{b}_x) \|_2 &\leq \alpha R, \label{cyl_con_1}\\ 
    -\alpha \frac{L}{2} \leq \gamma &\leq \alpha \frac{L}{2},\label{cyl_con_2}
\end{align}
where $\hat{b}_x = Q [1,0,0]^T$, and $\gamma \in \R{}$ is a slack variable.  These constraints contain a linear inequality and one second-order cone constraint, shown here in standard form:
\begin{align}
    \begin{bmatrix} 0 \\ 0 \end{bmatrix} - \begin{bmatrix} 0_{1 \times 3} & -L/2 & 1 \\ 0_{1 \times 3} & -L/2 & -1 \end{bmatrix} \begin{bmatrix} x \\ \alpha \\ \gamma \end{bmatrix} &\in \mathbf{R}_+^2, \label{eq:std_caps1}\\ 
    \begin{bmatrix}0 \\ -r  \end{bmatrix}  - \begin{bmatrix} 0_{1 \times 3} & -R & 0 \\ -I_3 & 0_{3 \times 1} & \hat{b}_x  \end{bmatrix} \begin{bmatrix} x \\ \alpha \\ \gamma \end{bmatrix} &\in \mathcal{Q}_4. \label{eq:std_caps2}
\end{align}
\subsubsection{Cylinder}
The description of a cylinder is shown in Fig. \ref{fig_cyl}, with an orientation, a radius $R$, and a length $L$. The constraints for this primitive are the same as for the capsule in equations \eqref{cyl_con_1} and \eqref{cyl_con_2}, with the introduction of two new scaled halfspace constraints that give the cylinder its flat ends:
\begin{align}
    [x - (r - \alpha \frac{L}{2} \hat{b}_x)]^T \hat{b}_x &\geq 0, \\ 
    [x - (r + \alpha \frac{L}{2} \hat{b}_x)]^T \hat{b}_x &\leq 0.
\end{align}
These constraints in standard form include those shown in equations \eqref{eq:std_caps1} and \eqref{eq:std_caps2} with the following addition:
\begin{align}
    \begin{bmatrix}  -\hat{b}_x^Tr \\ \phantom{-}\hat{b}_x^Tr \end{bmatrix} - \begin{bmatrix}  -\hat{b}_x^T & -L/2 & 0 \\ \phantom{-} \hat{b}_x^T & -L/2 & 0\end{bmatrix} \begin{bmatrix} x \\ \alpha \\ \gamma \end{bmatrix} &\in \mathbf{R}_+^2.
\end{align}
\subsubsection{Cone}
As shown in Fig. \ref{fig_cone}, a cone can be described with a height $H$, and a half angle $\beta$. The origin of the object-fixed frame $r$ is one-quarter of the way from the flat face to the point of the cone, and $\alpha$ scales the distance of these two ends from the center point $r$:
\begin{align}
    \|\tilde{x}_{2:3}\|_2 &\leq \tan(\beta) \tilde{x}_1, \\ 
    (x - r - \alpha \frac{H}{4} \hat{b}_x)^T \hat{b}_x &\leq 0 ,
\end{align}
where $\tilde{x} = Q^T  (x - r + \alpha \frac{3H}{4} \hat{b}_x)$.  These constraints in standard form are:
\begin{align}
    \hat{b}_x^Tr - \begin{bmatrix} \hat{b}_x^T  & -H/4  \end{bmatrix} \begin{bmatrix} x \\ \alpha \end{bmatrix} &\in \mathbf{R}_+^m, \\ 
    -EQ^Tr  - \begin{bmatrix} -EQ^T & v \end{bmatrix} \begin{bmatrix} x \\ \alpha \\  \end{bmatrix} &\in \mathcal{Q}_3,
\end{align}
where $E = \operatorname{diag}(\tan \beta,1,1)$ and $v = (-\frac{3H}{4} \tan \beta,0,0)$.
\subsubsection{Ellipsoid}
An ellipsoid, shown in Fig. \ref{fig_sphere}, can be described by a quadratic inequality $x^TPx \leq 1$, where $P\in \mathbf{S}^n_{++}$ is strictly positive definite and has an upper-triangular Cholesky factor $U \in \R{n \times n}$ \cite{boyd2004}. From here, a scaled ellipsoid with arbitrary position and orientation can be expressed in the following way:
\begin{align}
    \|U Q^T (x - r)\|_2 \leq \alpha,
\end{align}
where a sphere of radius $R$ is just a special case of an ellipsoid with $P = I/R^2$. These constraints can be written in standard form as:
\begin{align}
   \begin{bmatrix} 0 \\ -UQ^Tr \end{bmatrix} - \begin{bmatrix} 0_{1 \times 3} & -1 \\ -UQ^T & 0_{3 \times 1} \end{bmatrix} \begin{bmatrix} x \\ \alpha \end{bmatrix} &\in \mathcal{Q}_4 .
\end{align}
\subsubsection{Padded Polygon}
A ``padded" polygon is defined as the set of points within some radius $R$ of a two-dimensional polygon. Shown in Fig. \ref{fig_polyg}, the first two basis vectors of $\mathbb{B}$ span the polygon, and the polygon itself is defined with a slack variable $y \in \R{2}$, and $Cy\leq \alpha d$, where $C \in \R{m \times 2}$, and $d \in \R{m}$, describe the $m$ halfspace constraints for the polygon. This polygon is scaled in the same fashion as the polytope, and results in the following constraints:
\begin{align}
    \| x - (r + \tilde{Q} y)\|_2 &\leq \alpha R, \\ 
    Cy &\leq \alpha d,
\end{align}
where $\tilde{Q} \in \R{3\times 2}$ is the first two columns of $Q$. These constraints can be represented in standard form as the following:
\begin{align}
    0_{m} - \begin{bmatrix} 0_{m \times 3} & -d & C  \end{bmatrix} \begin{bmatrix} x \\ \alpha \\ y\end{bmatrix} &\in \mathbf{R}_+^m, \\
   \begin{bmatrix} 0 \\ -r \end{bmatrix} - \begin{bmatrix} 0_{1 \times 3} & -R & 0_{1 \times 2} \\ -I_3 & 0_{3 \times 1} & \tilde{Q} \end{bmatrix} \begin{bmatrix} x \\ \alpha \end{bmatrix} &\in \mathcal{Q}_4.
\end{align}
\subsection{Contact Points and Minimum Distance}
While the computation of contact points and minimum distance between primitives is not needed for any of the examples in Section \ref{sect:examples}, they are easy to compute with DCOL if desired. The intersection point on the two scaled primitives is referred to as $x^*$, but unless $\alpha^*=1$, this point does not exist on the surface of the primitives. The corresponding contact point for primitive $i$, $p_i \in \R{3}$, is calculated using the optimal $x^*$ and $\alpha^*$ from \eqref{dcd} as
\begin{align}
    p_i &= r_i + \frac{x^* - r_i}{\alpha^*},
\end{align}
where the intersection point between the scaled primitives is simply scaled back to each unscaled primitive. The distance between these points can also be calculated as follows:
\begin{align}
    \|d\|_2 = \|p_1 - p_2\|_2 &= \|r_1 - r_2 + \frac{r_2 - r_1}{\alpha}\|_2.
\end{align}
Both of these operations are fully differentiable given the derivatives from DCOL, allowing for the calculation of the sensitivities of the contact points with respect to the configurations of the primitives. 
\begin{figure*}[t!]
    \centering
    \includegraphics[width=.9\textwidth]{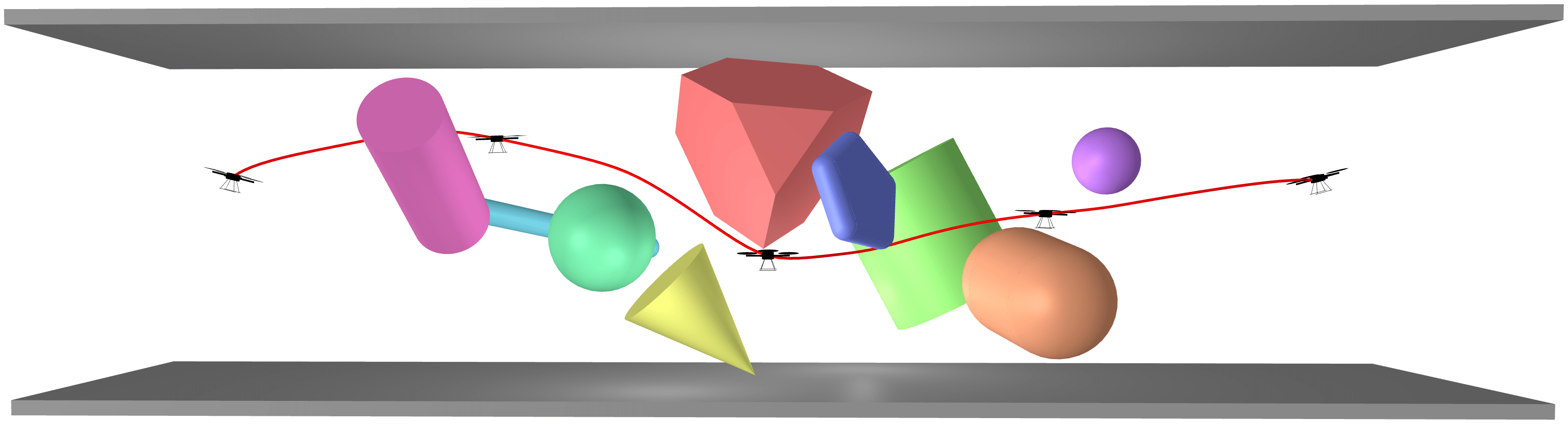}
    \caption{Trajectory optimization for a 6-DOF quadrotor as it moves from left to right through a cluttered hallway. The collision constraints were represented with DCOL, and the trajectory optimizer was initialized with a static hover at the initial condition.}
    \label{fig:hallway}
\end{figure*}
\section{Examples}\label{sect:examples}
In this section, we demonstrate the utility of differentiable collision detection in trajectory optimization problems where contact is to be avoided, and in physics simulation with contact where exact and differentiable collision information is required.  In both of these applications, a collision constraint $\alpha \geq 1$ is used to enforce no interpenetration between each pair of primitives, where $\alpha$ is the minimum scaling from DCOL.
\subsection{Trajectory Optimization}
Trajectory optimization is a powerful tool in motion planning and control, where a numerical optimization problem is formulated to solve for a constrained trajectory that minimizes a cost function.  A generic trajectory optimization problem with collision avoidance constraints from DCOL is as follows:
 \begin{equation}
	\begin{array}{ll}
	\underset{x_{1:N},u_{1:N-1}}{\mbox{minimize }} & \ell_N(x_N) + \sum_{k=1}^{N-1} \ell_k(x_k,u_k) \\
	\mbox{subject to } & x_{k+1} = f_k(x_k,u_k), \\
	                   & h_k(x_k,u_k) \leq 0, \\
                          & g_k(x_k,u_k) = 0, \\
                          & \alpha_k(x_k) \geq 1,
	\end{array} \label{eq:trajopt}
\end{equation}
where $k$ is the time step, $x_k$ and $u_k$ are the state and control inputs, $\ell_k$ and $\ell_N$ are the stage and terminal costs, $f(x_k,u_k)$ is the discrete dynamics function, $h_k(x_k,u_k)$ and $g_k(x_k,u_k)$ are inequality and equality constraints, and $\alpha_k(x_k)$ are the collision avoidance constraints from DCOL. Problems of this form can be solved with general purpose nonlinear program solvers like SNOPT \cite{gill2005}, and Ipopt \cite{wachter2006}, or more specialized solvers like ALTRO \cite{howell2019a,jackson2021c}. 

A key requirement for any gradient-based solver used to solve \eqref{eq:trajopt} is the ability to differentiate all of the cost and constraint functions with respect to the state and control inputs. This requirement has made collision-avoidance constraints difficult to incorporate into trajectory optimization frameworks because traditional collision detection methods are non-differentiable. In this section, DCOL is used to formulate collision-avoidance constraints in trajectory optimization problems to solve for collision-free trajectories.
\subsubsection{``Piano Mover'' Problem}
\begin{figure}[t]
\centerline{\includegraphics[width = .95\columnwidth]{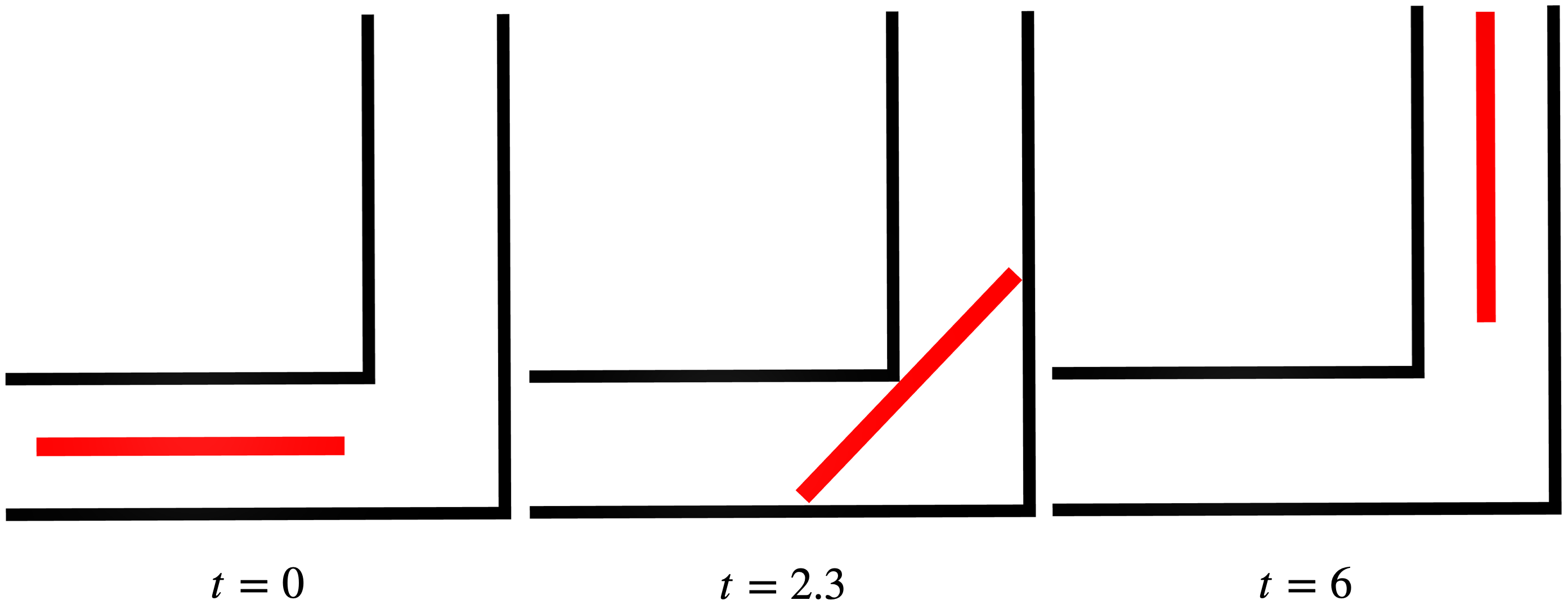}}
\caption{The \textit{``Piano Movers"} problem, where a ``piano'' (red rectangle) has to make a turn down a hallway, is solved with trajectory optimization. The piano and the walls are modeled as rectangular prisms. DCOL was used to represent all of the collision avoidance constraints that ensure the piano cannot travel through the wall, and the trajectory optimizer was able to converge on a feasible trajectory to deliver the piano to the goal state.}
\label{fig:piano}
\vspace{-10pt}
\end{figure}

The first problem we will look at is a variant of the \textit{``Piano Mover''} problem, where a piano must maneuver around a 90-degree turn in a hallway \cite{wilson2013,schwartz1983}. The walls are 1 meter apart, and the ``piano'' (a line segment) is 2.6 meters long, making the path around the corner nontrivial. This problem is solved with trajectory optimization and collision constraints, where the piano is parameterized as a cylindrical rigid body in two dimensions, with a position and orientation, and the hallway is modeled with polytopes. The solution to this problem is shown in Fig. \ref{fig:piano}, where the piano successfully maneuvers around the tight corner and reaches the goal state without traveling through any of the walls. The trajectory optimizer was initialized with the piano in a static pose at the initial condition.
\cite{wilson2013} \cite{schwartz1983}
\subsubsection{Quadrotor}
Motion planning for quadrotors has received significant attention in recent years \cite{mellinger2011,mellinger,sun2022}, with collision avoidance featured in many of these works \cite{falanga2020,penicka2022,shraim2018}. With DCOL, we are able to directly and exactly incorporate collision avoidance constraints into a quadrotor motion planner to solve for trajectories through cluttered environments. In this example, we use trajectory optimization for a classic 6-DOF quadrotor model from \cite{mellinger2011,jackson2020} to solve for a trajectory that traverses a cluttered hallway with 12 objects in it, shown in Fig. \ref{fig:hallway}. The solver was initialized with the quadrotor hovering at the initial condition, and a spherical outer approximation of the quadrotor geometry was used to compute collisions. Despite this naive guess, the solver was able to quickly converge on a collision-free trajectory through the obstacles. 
\subsubsection{Cone Through an Opening}
\begin{figure}
\centering
\subfloat[]{%
  \includegraphics[clip,width=.65\columnwidth]{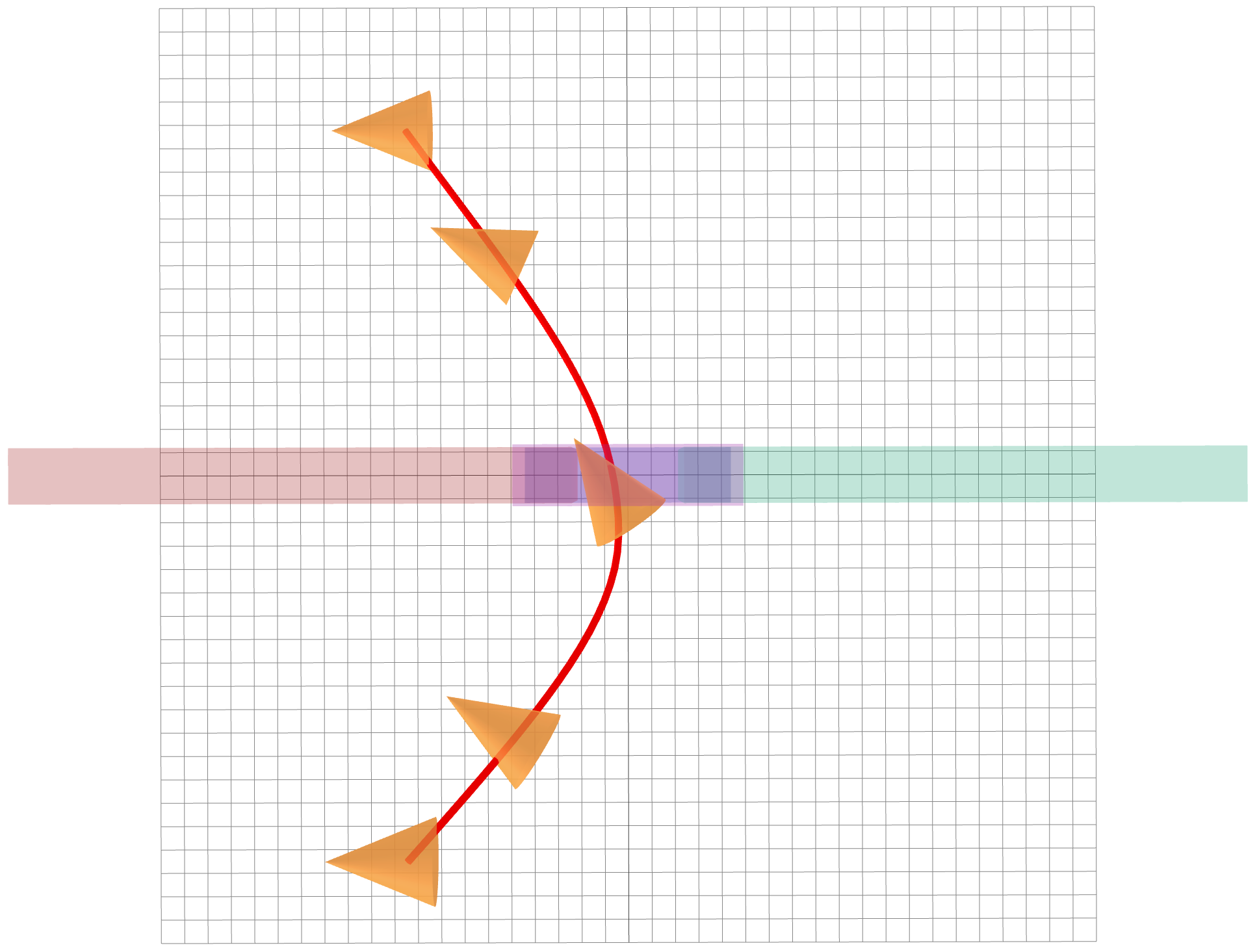}%
}
 
\subfloat[]{%
  \includegraphics[clip,width=0.8\columnwidth]{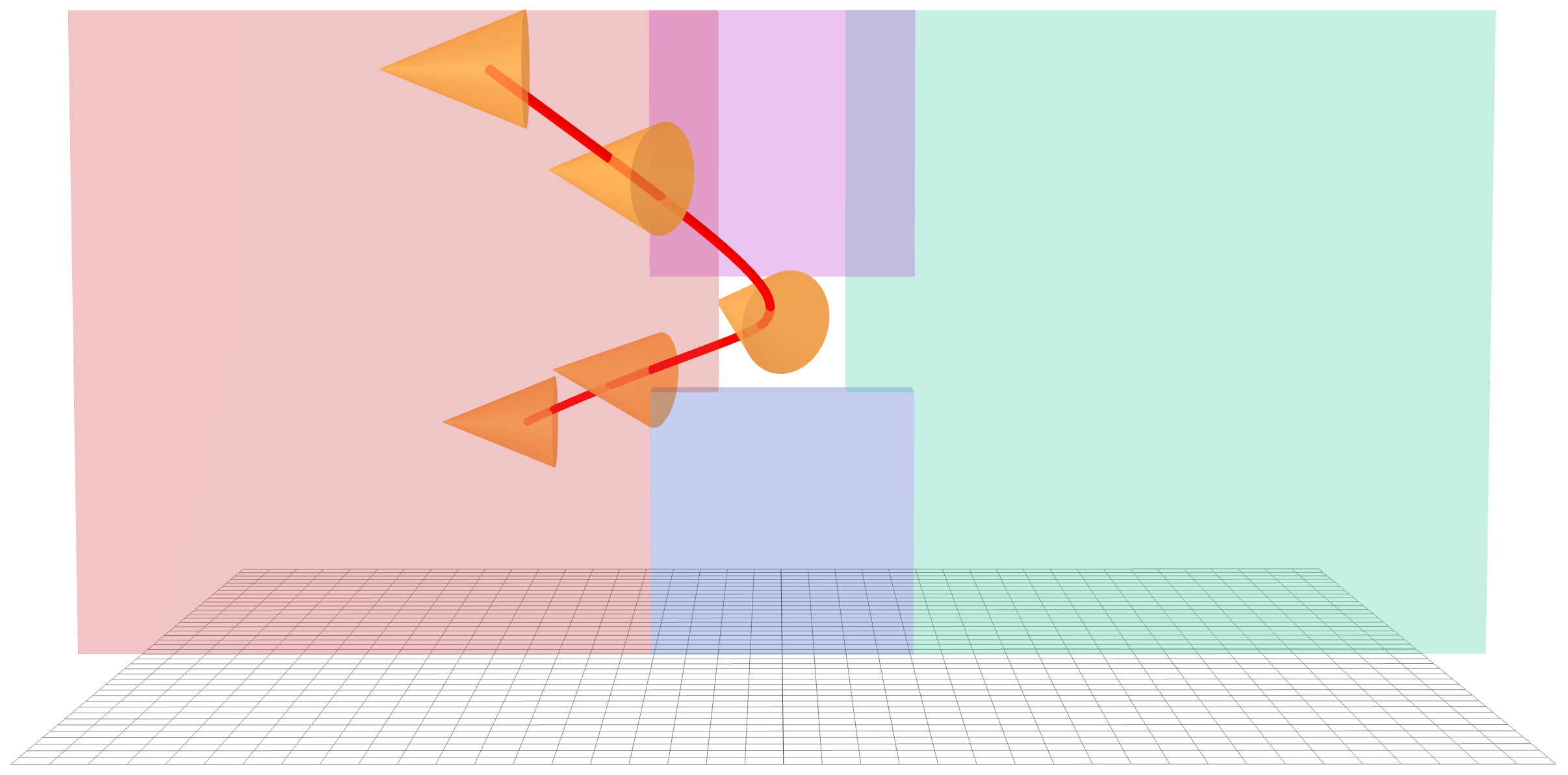}%
}
\caption{Trajectory optimization for a cone (orange) with translation and attitude control as it travels through a square opening in a wall. Top-down and side views are shown in (a) and (b), respectively. The cone is forced to slew to an attitude that allows for the passing of the cone through the opening before returning to the initial attitude. The trajectory optimizer was simply initialized with the static initial condition.}
\label{fig:conewall}
\end{figure}

This example demonstrates how trajectory optimization with DCOL can route a cone through a square hole in a wall, as shown in Fig. \ref{fig:conewall}. The dynamics of the cone are modeled as a rigid body with full translational and rotational control, and the wall is comprised of four rectangular prisms, making a rectangular opening in the wall. The trajectory optimizer converged on a solution where the cone successfully passes through the opening in the wall, requiring that the cone slewed its orientation and ``squeezed through'' the opening.  This example demonstrates the importance of the differentiability of the collision avoidance constraints, as the optimizer was forced to leverage both translational and rotational manipulation of the cone in order to successfully pass through the opening.  As with the previous two examples, there was no expert initial guess provided to the trajectory optimizer, just a static initial condition. 
\subsection{Contact Physics}
\begin{figure}[t]
\centerline{\includegraphics[width = .98\columnwidth]{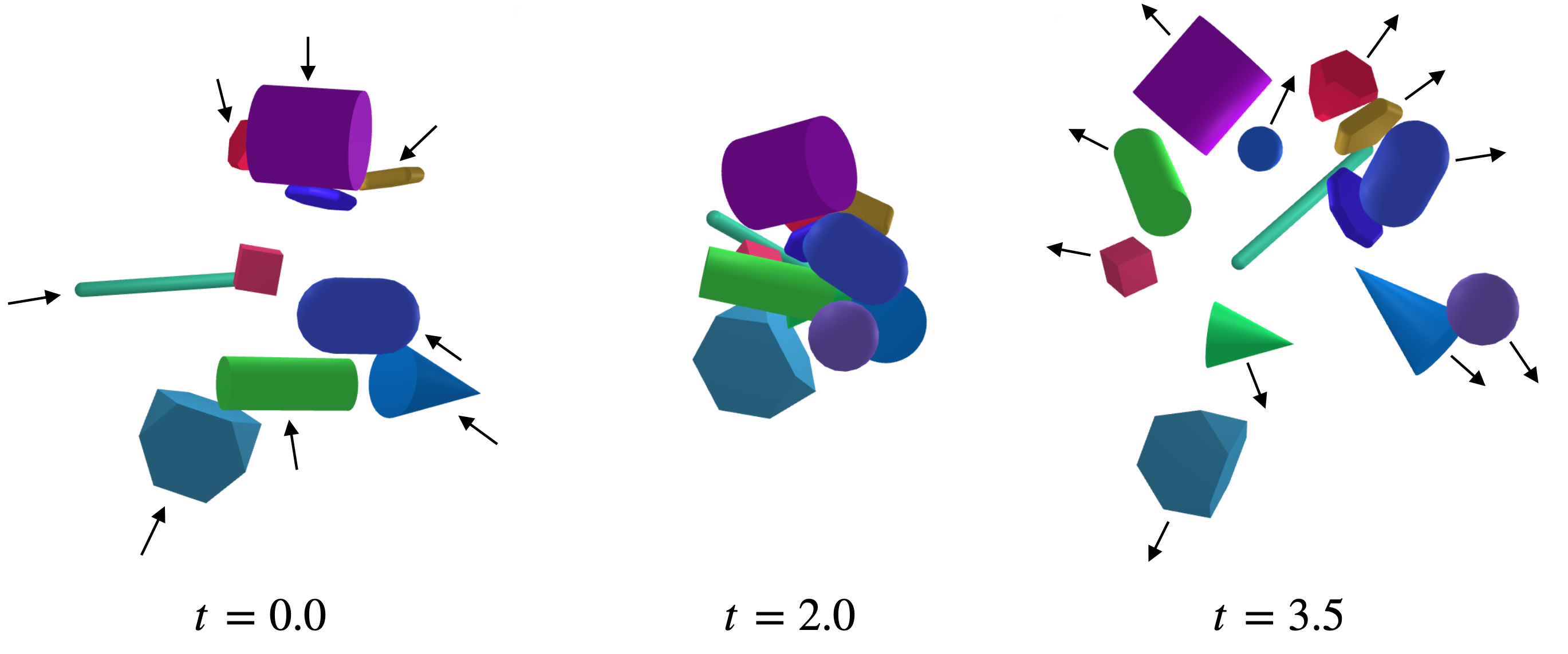}}
\caption{Contact physics with differentiable collision constraints embedded in a complementarity-based time-stepping scheme, simulated at 100 Hz.  Twelve convex objects are started at random positions with velocities pointing towards the origin at $t=0$. The objects impact each other at $t=2$ and spread out again by $t=3.5$.  Despite the complexity of the simulation, the collision constraints can be enforced to machine precision and the integration is stable.}
\label{fig:mashup}
\vspace{-10pt}
\end{figure}

Another application of differentiable collision detection in robotics is contact physics for simulation. Rigid-body mechanics with inelastic collisions can be simulated using complementarity-based time-stepping schemes \cite{howell2022}, where stationary points of a discretized action integral are solved for subject to contact constraints \cite{marsden2001}.  Normally these constraints are limited to traditionally differentiable ones like those between fixed contact points and a floor.  The differentiability of DCOL enables these same methods to be extended for simulating contact between convex primitives, as shown in Fig. \eqref{fig:mashup} where twelve primitives collide.  In terms of computation times, using DCOL for contact physics is reasonable given each constraint evaluation and differentiation are usually less than 10 $\mu$s as shown in Table \ref{speed_table}.
\section{Conclusion}\label{sect:conclusion}
We have presented DCOL, a fast differentiable collision detection algorithm capable of computing useful collision information and derivatives for pairs of any of six convex primitives. By formulating the collision-detection problem as an optimization problem that solves for the minimum uniform scaling that must be applied to each primitive before an intersection occurs, a surrogate proximity value is returned that is informative for primitives with or without a collision. Using differentiable convex optimization and a primal-dual interior-point conic solver, smooth derivatives of this optimization problem are returned after convergence with very little additional computation. The utility of DCOL is demonstrated in a wide variety of robotics applications, including motion planning and contact physics, where collision derivatives are required. Future work includes methods for convex decompositions of complex shapes as well as the incorporation of DCOL into existing physics engines. Our open-source Julia implementation of DCOL is available at \url{https://github.com/kevin-tracy/DifferentiableCollisions.jl}. 
\printbibliography

@article{agrawal2019,
  title = {Differentiating {{Through}} a {{Conic Program}}},
  author = {Agrawal, Akshay and Barratt, Shane and Boyd, Stephen and Moursi, Walaa M and Busseti, Enzo},
  year = {2019},
  journal = {Journal of Applied and Numerical Optimization},
  volume = {1},
  number = {2},
  pages = {107--115},
  langid = {english}
}

@article{agrawal2019a,
  title = {Differentiable {{Convex Optimization Layers}}},
  author = {Agrawal, Akshay and Amos, Brandon and Barratt, Shane and Boyd, Stephen and Diamond, Steven and Kolter, Zico},
  year = {2019},
  journal = {Advances in Neural Information Processing Systems},
  eprint = {1910.12430},
  eprinttype = {arxiv},
  pages = {9558--9570},
  archiveprefix = {arXiv}
}

@article{amos2019,
  title = {{{OptNet}}: {{Differentiable Optimization}} as a {{Layer}} in {{Neural Networks}}},
  shorttitle = {{{OptNet}}},
  author = {Amos, Brandon and Kolter, J. Zico},
  year = {2019},
  month = oct,
  journal = {arXiv:1703.00443 [cs, math, stat]},
  eprint = {1703.00443},
  eprinttype = {arxiv},
  primaryclass = {cs, math, stat},
  archiveprefix = {arXiv}
}

@article{andersen2003,
  title = {On Implementing a Primal-Dual Interior-Point Method for Conic Quadratic Optimization},
  author = {Andersen, E.D. and Roos, C. and Terlaky, T.},
  year = {2003},
  month = feb,
  journal = {Mathematical Programming},
  volume = {95},
  number = {2},
  pages = {249--277},
  issn = {0025-5610, 1436-4646},
  doi = {10.1007/s10107-002-0349-3},
  langid = {english}
}

@book{boyd2004,
  title = {Convex {{Optimization}}},
  author = {Boyd, Stephen and Vandenberghe, Lieven},
  year = {2004},
  publisher = {{Cambridge University Press}}
}

@inproceedings{cameron1997,
  title = {Enhancing {{GJK}}: Computing Minimum and Penetration Distances between Convex Polyhedra},
  shorttitle = {Enhancing {{GJK}}},
  booktitle = {Proceedings of {{International Conference}} on {{Robotics}} and {{Automation}}},
  author = {Cameron, S.},
  year = {1997},
  volume = {4},
  pages = {3112--3117},
  publisher = {{IEEE}},
  address = {{Albuquerque, NM, USA}},
  doi = {10.1109/ROBOT.1997.606761},
  isbn = {978-0-7803-3612-4},
  langid = {english}
}

@inproceedings{coumans2015,
  title = {Bullet {{Physics Simulation}}},
  booktitle = {{{SIGGRAPH}}},
  author = {Coumans, Erwin},
  year = {2015},
  publisher = {{ACM}},
  address = {{Los Angeles}},
  doi = {10.1145/2776880.2792704},
  isbn = {978-1-4503-3634-5}
}

@inproceedings{domahidi2013a,
  title = {{{ECOS}}: {{An SOCP}} Solver for Embedded Systems},
  shorttitle = {{{ECOS}}},
  booktitle = {2013 {{European Control Conference}} ({{ECC}})},
  author = {Domahidi, Alexander and Chu, Eric and Boyd, Stephen},
  year = {2013},
  month = jul,
  pages = {3071--3076},
  publisher = {{IEEE}},
  address = {{Zurich}},
  doi = {10.23919/ECC.2013.6669541},
  isbn = {978-3-033-03962-9},
  langid = {english}
}

@article{falanga2020,
  title = {Dynamic Obstacle Avoidance for Quadrotors with Event Cameras},
  author = {Falanga, Davide and Kleber, Kevin and Scaramuzza, Davide},
  year = {2020},
  month = mar,
  journal = {Science Robotics},
  volume = {5},
  number = {40},
  pages = {eaaz9712},
  publisher = {{American Association for the Advancement of Science}},
  doi = {10.1126/scirobotics.aaz9712}
}

@article{gilbert1988,
  title = {A Fast Procedure for Computing the Distance between Complex Objects in Three-Dimensional Space},
  author = {Gilbert, E.G. and Johnson, D.W. and Keerthi, S.S.},
  year = {1988},
  month = apr,
  journal = {IEEE Journal on Robotics and Automation},
  volume = {4},
  number = {2},
  pages = {193--203},
  issn = {08824967},
  doi = {10.1109/56.2083},
  langid = {english}
}

@inproceedings{gilbert1994,
  title = {New Distances for the Separation and Penetration of Objects},
  booktitle = {Proceedings of the 1994 {{IEEE International Conference}} on {{Robotics}} and {{Automation}}},
  author = {Gilbert, E.G. and {Chong Jin Ong}},
  year = {1994},
  pages = {579--586},
  publisher = {{IEEE Comput. Soc. Press}},
  address = {{San Diego, CA, USA}},
  doi = {10.1109/ROBOT.1994.351237},
  isbn = {978-0-8186-5330-8},
  langid = {english}
}

@article{gill2005,
  title = {{{SNOPT}}: {{An SQP Algorithm}} for {{Large-Scale Constrained Optimization}}},
  shorttitle = {{{SNOPT}}},
  author = {Gill, Philip E. and Murray, Walter and Saunders, Michael A.},
  year = {2005},
  month = jan,
  journal = {SIAM Review},
  volume = {47},
  number = {1},
  pages = {99--131},
  issn = {0036-1445, 1095-7200},
  doi = {10.1137/S0036144504446096},
  langid = {english}
}

@inproceedings{howell2019a,
  title = {{{ALTRO}}: {{A Fast Solver}} for {{Constrained Trajectory Optimization}}},
  booktitle = {{{IEEE}}/{{RSJ International Conference}} on {{Intelligent Robots}} and {{Systems}} ({{IROS}})},
  author = {Howell, Taylor A and Jackson, Brian E and Manchester, Zachary},
  year = {2019},
  month = nov,
  address = {{Macau, China}}
}

@article{howell2022,
  title = {Dojo: {{A Differentiable Physics Engine}} for {{Robotics}}},
  shorttitle = {Dojo},
  author = {Howell, Taylor A. and Cleac'h, Simon Le and Kolter, J. Zico and Schwager, Mac and Manchester, Zachary},
  year = {2022},
  month = jun,
  journal = {IEEE Transactions on Robotics and Automation (In Review)},
  doi = {10.48550/arXiv.2203.00806}
}

@inproceedings{jackson2020,
  title = {Scalable {{Cooperative Transport}} of {{Cable-Suspended Loads}} with {{UAVs}} Using {{Distributed Trajectory Optimization}}},
  booktitle = {International {{Conference}} on {{Robotics}} and {{Automation}}},
  author = {Jackson, Brian and Howell, Taylor and Shah, Kunal and Schwager, Mac and Manchester, Zachary},
  year = {2020},
  month = jun,
  pages = {8},
  address = {{Paris, France}}
}

@inproceedings{jackson2021c,
  title = {{{ALTRO-C}}: {{A Fast Solver}} for {{Conic Model-Predictive Control}}},
  booktitle = {International {{Conference}} on {{Robotics}} and {{Automation}} ({{ICRA}})},
  author = {Jackson, Brian E and Punnoose, Tarun and Neamati, Daniel and Tracy, Kevin and Jitosho, Rianna},
  year = {2021},
  pages = {8},
  address = {{Xi'an, China}}
}

@article{johnson,
  title = {Adaptive {{Trajectory Control}} for {{Autonomous Helicopters}}},
  author = {Johnson, Eric N and Kannan, Suresh K},
  volume = {28},
  number = {3},
  pages = {524--538},
  issn = {0731-5090, 1533-3884},
  doi = {10.2514/1.6271}
}

@article{lee2018,
  title = {{{DART}}: {{Dynamic Animation}} and {{Robotics Toolkit}}},
  shorttitle = {{{DART}}},
  author = {Lee, Jeongseok and X. Grey, Michael and Ha, Sehoon and Kunz, Tobias and Jain, Sumit and Ye, Yuting and S. Srinivasa, Siddhartha and Stilman, Mike and Karen Liu, C.},
  year = {2018},
  month = feb,
  journal = {The Journal of Open Source Software},
  volume = {3},
  number = {22},
  pages = {500},
  issn = {2475-9066},
  doi = {10.21105/joss.00500},
  langid = {english}
}

@article{marsden2001,
  title = {Discrete {{Mechanics}} and {{Variational Integrators}}},
  author = {Marsden, J. E. and West, M.},
  year = {2001},
  journal = {Acta Numerica},
  volume = {10},
  pages = {357--514}
}

@article{mellinger,
  title = {Trajectory {{Generation}} and {{Control}} for {{Precise Aggressive Maneuvers}} with {{Quadrotors}}},
  author = {Mellinger, Daniel and Michael, Nathan and Kumar, Vijay},
  pages = {13},
  langid = {english}
}

@inproceedings{mellinger2011,
  title = {Minimum Snap Trajectory Generation and Control for Quadrotors},
  booktitle = {2011 {{IEEE International Conference}} on {{Robotics}} and {{Automation}} ({{ICRA}})},
  author = {Mellinger, D. and Kumar, V.},
  year = {2011},
  month = may,
  pages = {2520--2525},
  doi = {10.1109/ICRA.2011.5980409}
}

@misc{montaut2022a,
  title = {Differentiable {{Collision Detection}}: A {{Randomized Smoothing Approach}}},
  shorttitle = {Differentiable {{Collision Detection}}},
  author = {Montaut, Louis and Lidec, Quentin Le and Bambade, Antoine and Petrik, Vladimir and Sivic, Josef and Carpentier, Justin},
  year = {2022},
  month = sep,
  number = {arXiv:2209.09012},
  eprint = {2209.09012},
  eprinttype = {arxiv},
  primaryclass = {cs},
  publisher = {{arXiv}},
  archiveprefix = {arXiv},
  langid = {english}
}

@techreport{mosekaps2014,
  title = {The {{MOSEK}} Optimization Software},
  author = {{Mosek ApS}},
  year = {2014}
}

@article{nesterov1997,
  title = {Self-{{Scaled Barriers}} and {{Interior-Point Methods}} for {{Convex Programming}}},
  author = {Nesterov, Yu. E. and Todd, M. J.},
  year = {1997},
  month = feb,
  journal = {Mathematics of Operations Research},
  volume = {22},
  number = {1},
  pages = {1--42},
  issn = {0364-765X, 1526-5471},
  doi = {10.1287/moor.22.1.1},
  langid = {english}
}

@article{nesterov1998,
  title = {Primal-{{Dual Interior-Point Methods}} for {{Self-Scaled Cones}}},
  author = {Nesterov, Yu. E. and Todd, M. J.},
  year = {1998},
  month = may,
  journal = {SIAM Journal on Optimization},
  volume = {8},
  number = {2},
  pages = {324--364},
  issn = {1052-6234, 1095-7189},
  doi = {10.1137/S1052623495290209},
  langid = {english}
}

@phdthesis{newth2013,
  type = {Master of {{Science}}},
  title = {Minkowski {{Portal Refinement}} and {{Speculative Contacts}} in {{Box2D}}},
  author = {Newth, Joshua},
  year = {2013},
  month = apr,
  address = {{San Jose, CA, USA}},
  doi = {10.31979/etd.q6rm-ch9a},
  langid = {english},
  school = {San Jose State University}
}

@inproceedings{pan2012,
  title = {{{FCL}}: {{A}} General Purpose Library for Collision and Proximity Queries},
  shorttitle = {{{FCL}}},
  booktitle = {2012 {{IEEE International Conference}} on {{Robotics}} and {{Automation}}},
  author = {Pan, Jia and Chitta, Sachin and Manocha, Dinesh},
  year = {2012},
  month = may,
  pages = {3859--3866},
  publisher = {{IEEE}},
  address = {{St Paul, MN, USA}},
  doi = {10.1109/ICRA.2012.6225337},
  isbn = {978-1-4673-1405-3},
  langid = {english}
}

@article{penicka2022,
  title = {Learning {{Minimum-Time Flight}} in {{Cluttered Environments}}},
  author = {Penicka, Robert and Song, Yunlong and Kaufmann, Elia and Scaramuzza, Davide},
  year = {2022},
  month = jul,
  journal = {IEEE Robotics and Automation Letters},
  volume = {7},
  number = {3},
  pages = {7209--7216},
  issn = {2377-3766, 2377-3774},
  doi = {10.1109/LRA.2022.3181755},
  langid = {english}
}

@article{schwartz1983,
  title = {On the ``Piano Movers''' Problem {{I}}. {{The}} Case of a Two-Dimensional Rigid Polygonal Body Moving amidst Polygonal Barriers},
  shorttitle = {On the ``Piano Movers''' Problem {{I}}. {{The}} Case of a Two-Dimensional Rigid Polygonal Body Moving amidst Polygonal Barriers},
  author = {Schwartz, Jacob T. and Sharir, Micha},
  year = {1983},
  month = may,
  journal = {Communications on Pure and Applied Mathematics},
  volume = {36},
  number = {3},
  pages = {345--398},
  issn = {00103640},
  doi = {10.1002/cpa.3160360305},
  langid = {english}
}

@article{shraim2018,
  title = {A Survey on Quadrotors: {{Configurations}}, Modeling and Identification, Control, Collision Avoidance, Fault Diagnosis and Tolerant Control},
  shorttitle = {A Survey on Quadrotors},
  author = {Shraim, Hassan and Awada, Ali and Youness, Rafic},
  year = {2018},
  month = jul,
  journal = {IEEE Aerospace and Electronic Systems Magazine},
  volume = {33},
  number = {7},
  pages = {14--33},
  issn = {0885-8985, 1557-959X},
  doi = {10.1109/MAES.2018.160246},
  langid = {english}
}

@article{snethen2008,
  title = {{{XenoCollide}}: {{Complex Collision Made Simple}}},
  shorttitle = {{{XenoCollide}}},
  author = {Snethen, Gary},
  year = {2008},
  journal = {undefined},
  langid = {english}
}

@article{sun2022,
  title = {A {{Comparative Study}} of {{Nonlinear MPC}} and {{Differential-Flatness-Based Control}} for {{Quadrotor Agile Flight}}},
  author = {Sun, Sihao and Romero, Angel and Foehn, Philipp and Kaufmann, Elia and Scaramuzza, Davide},
  year = {2022},
  journal = {IEEE Transactions on Robotics},
  pages = {1--17},
  issn = {1552-3098, 1941-0468},
  doi = {10.1109/TRO.2022.3177279},
  langid = {english}
}

@article{tedrake2019a,
  title = {Drake: {{Model-based}} Design and Verification for Robotics},
  author = {Tedrake, Russ and {The Drake Development Team}},
  year = {2019}
}

@inproceedings{todorov2012a,
  title = {{{MuJoCo}}: {{A}} Physics Engine for Model-Based Control},
  booktitle = {2012 {{IEEE}}/{{RSJ International Conference}} on {{Intelligent Robots}} and {{Systems}}},
  author = {Todorov, Emanuel and Erez, Tom and Tassa, Yuval},
  year = {2012},
  pages = {5026--5033},
  doi = {10.1109/IROS.2012.6386109}
}

@article{tracy2022,
  title = {{{DiffPills}}: {{Differentiable Collision Detection}} for {{Capsules}} and {{Padded Polygons}}},
  shorttitle = {{{DiffPills}}},
  author = {Tracy, Kevin and Howell, Taylor A. and Manchester, Zachary},
  year = {2022},
  month = jul,
  journal = {http://arxiv.org/abs/2207.00202},
  eprint = {2207.00202},
  eprinttype = {arxiv},
  doi = {10.48550/arXiv.2207.00202},
  archiveprefix = {arXiv}
}

@article{vandenbergen2001,
  title = {Proximity {{Queries}} and {{Penetration Depth Computation}} on {{3D GAme Objects}}},
  author = {Van Den Bergen, Gino},
  year = {2001}
}

@article{vandenberghe,
  title = {The {{CVXOPT}} Linear and Quadratic Cone Program Solvers},
  author = {Vandenberghe, L},
  pages = {30},
  langid = {english}
}

@article{wachter2006,
  title = {On the Implementation of an Interior-Point Filter Line-Search Algorithm for Large-Scale Nonlinear Programming},
  author = {W{\"a}chter, Andreas and Biegler, Lorenz T.},
  year = {2006},
  month = mar,
  journal = {Mathematical Programming},
  volume = {106},
  number = {1},
  pages = {25--57},
  issn = {0025-5610, 1436-4646},
  doi = {10.1007/s10107-004-0559-y},
  langid = {english}
}

@inproceedings{wilson2013,
  title = {A "{{Piano Movers}}" {{Problem Reformulated}}},
  booktitle = {2013 15th {{International Symposium}} on {{Symbolic}} and {{Numeric Algorithms}} for {{Scientific Computing}}},
  author = {Wilson, David and Davenport, James H. and England, Matthew and Bradford, Russell},
  year = {2013},
  month = sep,
  eprint = {1309.1588},
  eprinttype = {arxiv},
  primaryclass = {cs},
  pages = {53--60},
  doi = {10.1109/SYNASC.2013.14},
  archiveprefix = {arXiv},
  langid = {english}
}

@misc{zimmermann2022,
  title = {Differentiable {{Collision Avoidance Using Collision Primitives}}},
  author = {Zimmermann, Simon and Busenhart, Matthias and Huber, Simon and Poranne, Roi and Coros, Stelian},
  year = {2022},
  month = apr,
  number = {arXiv:2204.09352},
  eprint = {2204.09352},
  eprinttype = {arxiv},
  primaryclass = {cs},
  publisher = {{arXiv}},
  archiveprefix = {arXiv},
  langid = {english}
}
\vspace{12pt}
\end{document}